\documentclass[10pt,conference]{IEEEtran}
\IEEEoverridecommandlockouts

\usepackage{graphicx}
\usepackage{amsmath}
\usepackage{amssymb}
\usepackage{mathtools}
\usepackage{subcaption}
\usepackage{hyperref}
\usepackage[nolist,nohyperlinks]{acronym}
\usepackage{orcidlink}

\newcommand{\R}{\mathbb{R}}

\title{Exploring Sparsity and Smoothness of Arbitrary $\ell_p$ Norms in Adversarial Attacks}
\author{\IEEEauthorblockN{Christof Duhme \orcidlink{0009-0004-1853-2862}}
\IEEEauthorblockA{\textit{Department of Computer Science} \\
\textit{University of Münster}\\
Münster, Germany}
\and
\IEEEauthorblockN{Florian Eilers \orcidlink{0000-0003-0726-3287}}
\IEEEauthorblockA{\textit{Department of Computer Science} \\
\textit{University of Münster}\\
Münster, Germany}
\and
\IEEEauthorblockN{Xiaoyi Jiang \orcidlink{0000-0001-7678-9528}}
\IEEEauthorblockA{\textit{Department of Computer Science} \\
\textit{University of Münster}\\
Münster, Germany}}

\begin{document}

\begin{acronym}
    \acro{pgd}[PGD]{Projected Gradient Descent}
    \acro{apgd}[APGD]{Auto-Projected Gradient Descent}
    \acro{afw}[AFW]{Auto-Frank-Wolfe}
    \acro{gtsrb}[GTSRB]{German Traffic Sign Recognition Benchmark}
\end{acronym}

\maketitle

\begin{abstract}
Adversarial attacks against deep neural networks are commonly constructed under $\ell_p$ norm constraints, most often using $p=1$, $p=2$ or $p=\infty$, and potentially regularized for specific demands such as sparsity or smoothness.
These choices are typically made without a systematic investigation of how the norm parameter \( p \) influences the structural and perceptual properties of adversarial perturbations.
In this work, we study how the choice of \( p \) affects sparsity and smoothness of adversarial attacks generated under \( \ell_p \) norm constraints for values of $p \in [1,2]$. 
To enable a quantitative analysis, we adopt two established sparsity measures from the literature and introduce three smoothness measures.
In particular, we propose a general framework for deriving smoothness measures based on smoothing operations and additionally introduce a smoothness measure based on first-order Taylor approximations.
Using these measures, we conduct a comprehensive empirical evaluation across multiple real-world image datasets and a diverse set of model architectures, including both convolutional and transformer-based networks.
We show that the choice of $\ell_1$ or $\ell_2$ is suboptimal in most cases and the optimal $p$ value is dependent on the specific task.
In our experiments, using $\ell_p$ norms with $p\in [1.3, 1.5]$ yields the best trade-off between sparse and smooth attacks.
These findings highlight the importance of principled norm selection when designing and evaluating adversarial attacks.
\end{abstract}

\textbf{\textcopyright 2026 IEEE}. Personal use of this material is permitted.  Permission from IEEE must be obtained for all other uses, in any current or future media, including reprinting/republishing this material for advertising or promotional purposes, creating new collective works, for resale or redistribution to servers or lists, or reuse of any copyrighted component of this work in other works.

\section{Introduction}
Deep neural networks are successfully applied in many academic, corporate and public areas, including safety critical applications such as health care or autonomous driving.
In these settings, adversarial attacks pose a serious security threat \cite{zhou2021machine,zhang2022evaluating}, as small and often imperceptible perturbations to the input can lead to incorrect model predictions.
Understanding the structure and properties of adversarial attacks is therefore essential for designing models that are robust against such threats.

The general objective of adversarial attacks is to modify an input in a way that is difficult for humans to perceive while causing a model to change its prediction.
Adversarial perturbations are typically constrained to be small according to some norm.
In addition to the overall magnitude of the perturbation, both sparsity and smoothness have been shown to play an important role in the perceptual quality of adversarial examples \cite{heng2018harmonic,croce2019sparse}.

Adversarial attacks are often framed as an optimization problem
\begin{align}\label{eq:aa}
    \max_{||\delta||_p < \varepsilon} l(f(x + \delta), y)
\end{align}
where $f$ denotes a neural network, $x$ an input image with label $y$, $l$ a loss function, and $\varepsilon > 0$ controls the attack strength.

Many works have investigated how to regularize this optimization problem to promote smooth \cite{zhang2020smooth} or sparse \cite{su2019one,sadiku2023group} perturbations.
Other works have focused on restricting the search space from $\{||\delta||_p < \varepsilon\}$ to a search space that only allows smooth \cite{heng2018harmonic,guo2019low,liu2023ssta} or sparse \cite{croce2019sparse,liu2023low} solutions.
In these frameworks, the $\ell_p$ norm $||\cdot||_p$ is regularly defaulted to $p=1$, $p=2$ or $p=\infty$ without further investigation.

To the best of our knowledge, no work so far has explicitly investigated how the choice of $p$ effects the sparsity and smoothness of the adversarial attack.
In this work, we study the behavior of adversarial perturbations generated under $\ell_p$ norm constraints for values of $p \in [1, 2]$.
Our analysis focuses on quantifying how sparsity and smoothness change as functions of $p$ across different model architectures and datasets.

Sparsity in adversarial attacks is commonly understood as perturbing only a small number of pixels.
Accordingly, many approaches rely on the $\ell_0$ norm to enforce sparsity.
We adopt a broader notion of sparsity, in which only a small number of pixels in the adversarial attack change significantly but tiny changes in more pixels are neglected, which is not covered by the $\ell_0$ constraint.
This is in line with \cite{hurley2009comparing} who present desirable properties for sparsity measures and evaluate them on commonly used measures.
We choose the two sparsity measures investigated in their work that satisfy the most desirable properties.

To the best of our knowledge, no smoothness measures for adversarial attacks have been studied.
We introduce a general framework for construction smoothness measures based on smoothing operations.
In addition, we propose a smoothness measure derived from first-order Taylor approximations.
These measures allow us to quantitatively assess the smoothness of adversarial perturbations generated under different $\ell_p$ norm constraints with $p \in [1,2]$ in a white box scenario on different datasets and architectures.

To summarize, our contributions are as follows:
\begin{enumerate}
    \item We propose a general framework for deriving smoothness measures from smoothing operations.
    \item We introduce two smoothness measures based on this framework and additionally propose a smoothness measure based on first-order Taylor approximations.
    \item We analyze sparsity and smoothness of adversarial attacks generated under arbitrary $\ell_p$ norm constraints with these three smoothness measures and two well-known sparsity measures from literature.
    \item We show that neither of the standard choices $p=1$ and $p=2$ is optimal with regard to sparsity and smoothness, but the optimal choice $p$ depends on both the model architecture and the dataset.
\end{enumerate}

The work is structured as follows.
Section~\ref{sec:rel_work} reviews related work on sparsity and smoothness in adversarial attacks.
Section~\ref{sec:backround} provides methodological background and Section~\ref{sec:smooth} introduces the proposed smoothness measures.
Section~\ref{sec:experiments} describes the experimental setup, followed by the presentation and analysis of the experimental results in Section~\ref{sec:results}.
Finally, Section~\ref{sec:conclusion} concludes the paper.

\section{Related Work}\label{sec:rel_work}
Ever since the introduction of adversarial examples \cite{szegedy2014intriguing}, extensive research has explored attack and defense methods in computer vision, with recent surveys summarizing a decade of progress across vision tasks and reviewing domain-specific challenges in medical image analysis \cite{Dong2025, Zhang2026}. In contrast to these broad overviews, we focus on how varying $p$ in $\ell_p$-constrained attacks affects the structural properties of perturbations, particularly sparsity and smoothness.

Most attacks are based on approximately solving the optimization problem in \eqref{eq:aa} using gradient-based methods. 
But black-box attacks \cite{wang2025greedypixel} and generative models have also been used to produce adversarial perturbations, including GAN-based approaches \cite{he2022transferable} and diffusion-based methods \cite{xue2023diffusion,chen2023advdiffuser}.
In addition, adversarial attacks in the physical world have been studied in a variety of settings \cite{kurakin2018adversarial,ren2021adversarial}.

\paragraph{Optimization-based attacks and evaluation}
To solve \eqref{eq:aa} with gradient descent, \ac{pgd} was introduced \cite{madry2018towards} and later enhanced to the parameter-free \ac{apgd} \cite{croce2020reliable}.
Croce and Hein subsequently extended this line of work to $\ell_1$-\ac{apgd} \cite{croce2021mind}, which explicitly respects image-domain constraints (e.g., $x+\delta \in [0,1]^n$) and was designed to produce sparse perturbations.
However, they did not verify this sparsity claim with independent measure.
For attacks under more general $\ell_p$ constraints, Frank-Wolfe \cite{frank1956algorithm} based approaches provide an alternative to projected methods by avoiding explicit projections \cite{chen2020frank}, but only focus on $p=\infty$ in the evaluation.
Building on this direction, Boreiko et al.\ introduced \ac{afw} as an adaptive Frank-Wolfe scheme to generate perturbations under arbitrary $\ell_p$ constraints for $p>1$ \cite{boreiko2022sparse}.
Recent work explores optimization strategies for sparsity beyond classical norms. For example, Cinà et al.\ proposed gradient-based optimization of generalized zero-norm attacks evaluating relaxations of sparsity measures \cite{cina2025sigma}, illustrating that alternative norm relaxations yield improved empirical trade-offs between accuracy and perturbation size.

\paragraph{Norm choice and perceptual relevance}
Despite the widespread use of pixel-space $\ell_p$ norms (most commonly $p \in \{1,2,\infty\}$), the relationship between these norms and human perception is not straightforward.
Sen et al.\ explicitly ask whether adversarial attacks should be evaluated using pixel $p$-norms and discuss alternative similarity measures \cite{sen2019should}.
Relatedly, perceptual threat models and attacks based on learned perceptual distances have been explored, for example via neural perceptual threat models \cite{laidlaw2020perceptual}, and by considering perturbation classes that can be perceptually realistic despite large pixel $\ell_p$ distances, such as spatial transformations \cite{xiao2018spatially}.
However, recent meta-attack frameworks such as DAASH demonstrate that combining multiple `\(\ell_p\)` based attacks with perceptual distortion metrics (e.g., SSIM, LPIPS) yields perturbations with superior perceptual quality even when constrained by pixel norms \cite{nafi2025dash}.

\paragraph{Sparse adversarial attacks}
Sparsity and smoothness are often cited as desirable properties for imperceptible adversarial perturbations, motivating methods that explicitly target these characteristics.
To generate sparse attacks, Croce and Hein propose attacks minimizing $\ell_0$-distance and variants that additionally control per-pixel changes to improve imperceptibility \cite{croce2019sparse}.
Other approaches factorize the perturbation into magnitude and pixel selection components \cite{fan2020sparse}, learn distortion maps indicating less perceptually sensitive regions \cite{dong2020greedyfool}, or consider extreme settings such as one-pixel attacks \cite{su2019one}.

\paragraph{Smooth and frequency-structured attacks}
To promote smooth perturbations, Zhang et al.\ integrate Laplacian smoothing into the attack pipeline \cite{zhang2020smooth}.
Complementary approaches restrict the search space to low-frequency perturbations \cite{guo2019low} or generate perturbations within function classes that yield visually structured changes, such as harmonic functions \cite{heng2018harmonic}.
More recent approaches target structured sparsity: Lin et al.\ propose interpretable sparse attacks that minimize the number of perturbed pixels while preserving attack effectiveness \cite{lin2025towards}, and Heshmati et al.\ introduce ATOS, a framework that generates structured, group-sparse adversarial perturbations that highlight class-relevant features \cite{heshmati2025versatile}.

\paragraph{Closest prior work}
The most direct prior work to our study is Boreiko et al.\ \cite{boreiko2022sparse}, who generate sparse visual counterfactual explanations (VCEs) using an intermediate norm $\ell_{1.5}$ and argue qualitatively for a favorable trade-off between sparsity and smoothness.
However, their choice of $p$ is fixed and the effect of varying $p$ is not quantified.
In contrast, we systematically analyze how varying $p \in [1,2]$ affects sparsity and smoothness using multiple quantitative measures, and we provide a method to identify the optimal $p$ per model and dataset.

\section{Background}\label{sec:backround}
We will give a short introduction to adversarial attacks with arbitrary $\ell_p$ norm constraints as well as sparsity measures. 

\subsection{\texorpdfstring{$\ell_p$}{lp} Norms}\label{subsec:lpnorms}
As defined in \eqref{eq:aa}, adversarial attacks are commonly formulated as constrained optimization problem, where the search space for the perturbation $\delta$ is given by the $\ell_p$ ball $B^p_\varepsilon = \{ \delta \in \R^n | \ ||\delta ||_p < \varepsilon\}$ where $n$ denotes the image dimension and $\varepsilon>0$ controls the attack strength.
While most prior work focuses on specific choices $p=1$, $p=2$ or $p=\infty$, $p$ does not need to be restricted to these values.
For $1 \leq p < \infty$, the $\ell_p$ norm in $\R^n$ is defined as:
\begin{align}
    ||x||_p = \sum_{k=1}^n \left(|x_k|^p\right)^{\frac{1}{p}}
\end{align}

\subsection{Adversarial Attacks for Arbitrary \texorpdfstring{$\ell_p$}{lp} Norms}\label{subsec:aa_arbp}
\ac{apgd} \cite{croce2021mind} was introduced to eliminate the need for manual hyperparameter tuning in \ac{pgd} \cite{madry2018towards}.
While \ac{apgd} efficiently solves the optimization problem in \eqref{eq:aa}, it does not respect the image domain (usually $[0, 1]$) and relies on clipping after calculation of the attack.
To address this limitation for sparse attacks, \ac{apgd} was extended to $\ell_1$-\ac{apgd} \cite{croce2021mind}, which explicitly respects image-domain constraints while solving the adversarial optimization problem for the $\ell_1$ norm.

Both \ac{apgd} and $\ell_1$-\ac{apgd} are restricted to specific norm choices and cannot be directly applied to arbitrary values of $p$.
Therefore, \ac{afw} \cite{boreiko2022sparse} was introduced as an adaptive version of the Frank-Wolfe algorithm \cite{chen2020frank} that can iteratively solve the optimization problem for arbitrary $\ell_p$ norms for $p > 1$, while explicitly respecting the image domain.

Let $w \in \mathbb{R}^n, \ w = \nabla_x l(f(x), y)$ be the gradient. 
To avoid the necessary projection step as required to solve \eqref{eq:aa} by \ac{apgd} Boreiko et al. then reformulate the optimization problem as:
\begin{align}\label{eq:afw}
    \underset{\delta \in \mathbb{R}^n}{\arg \max} \langle w, \delta \rangle \; \text{s.t.} \; ||\delta||_p \leq \varepsilon, \; x+\delta \in [0,1]^n
\end{align}
The closed-form solution of this optimization step is given by
\begin{align}
    \delta_i^* = \min \left\{ \gamma_i, \left( \frac{|w_i|}{p \mu^*} \right) \right\} \text{sign} w_i
\end{align}
where
\begin{align}
\gamma_i = \max \{ -x_i \text{ sign} (w_i), (1 - x_i) \text{ sign} (w_i) \}
\end{align}
and $\mu^* > 0$, which can be computed in $O(n \log n)$ time.

The solution \eqref{eq:afw} provides a framework to calculate adversarial attacks for $p \ge 1$ while respecting the image constraints $[0, 1]$.
We restrict our empirical analysis to $p \in [1,2]$.

\subsection{Sparsity Measures}\label{subsec:sparse}
Sparsity is commonly cited as a desirable property of adversarial perturbations, as perturbing only a small number of pixels is often associated with increased imperceptibility.
It is often just measured by the $\ell_0$ measure
\begin{align}
    \ell_0(x) = |\{x_i \neq 0\}|
\end{align}
We argue, however, that small perturbations are imperceptible anyways, thus a less strict notion of sparsity that accounts for the distribution of perturbation magnitudes is more informative in the setting of adversarial attacks.
Following prior work \cite{hurley2009comparing}, we adopt sparsity measures that characterize how energy is distributed across coefficients.
In this case, the coefficients are the adversarial perturbations.
Such measures have desirable properties originally studied in economic settings \cite{dalton1920measurement,rickard2004gini}, where sparsity is analogous to wealth concentration.

We choose the Gini Index and the Hoyer measure, since they satisfy the most important properties for fixed-size inputs such as images.

The Gini Index \cite{gini1921measurement} is defined as
\begin{align}
    S_\text{Gini}(c) = 1 - 2 \sum_{k=1}^N \frac{c_k}{||c||_1} \left( \frac{N - k + \frac{1}{2}}{N} \right)
\end{align}
for ordered data $c_1 \leq c_2 \leq ... \leq c_N$.
The Gini Index is a weighted sum that remains sensitive to small coefficients and is normalized with respect to the number of coefficients, making it suitable for comparisons across datasets.

The Hoyer measure \cite{hoyer2004non} is defined as
\begin{align}
    S_\text{Hoyer}(c) = \left( \sqrt{N} - \frac{\sum_j c_j}{\sqrt{\sum_j c_j^2}} \right) \left( \sqrt{N} - 1 \right)^{-1}.
\end{align}
This is a normalized ratio between $\ell_1$ and $\ell_2$.
It is zero if and only if all coefficients are equal, and one if and only if exactly one coefficient is non-zero.

\section{Smoothness Measures}\label{sec:smooth}
To the best of our knowledge, there are no established quantitative smoothness measures specifically designed for adversarial perturbations.
We therefore introduce two complementary approaches to measure smoothness.
First, we propose a general framework that derives smoothness measures from smoothing operators.
Second, we introduce a smoothness measure based on first-order Taylor approximations.
Throughout this section, we interpret higher smoothness values as indicating smoother perturbations.

\subsection{Smoothing Operation based Smoothness Measure}
We propose a general framework for constructing smoothness measures based on smoothing operations.
The central observation underlying this framework is that a smooth signal changes less under the application of a smoothing operator than a non-smooth signal.
 
Let $I$ denote an image (or perturbation), and let $C_\alpha$ be a smoothing operator parameterized by a non-negative smoothing strength $\alpha$, where larger values of $\alpha$ correspond to stronger smoothing. We denote by $C_\alpha(I)$ the result of applying the smoothing operator $C_\alpha$ to $I$.
We define the $C$-smoothness $T_C$ of an image $I$ as a negative exponentially weighted average of the deviation between $I$ and its smoothed version $C_\alpha(I)$:
\begin{align}
    T_C(I) = -\int_{\alpha \in \R^+} \exp(-\alpha) \ |C_\alpha(I) - I| \ d\alpha
\end{align}
The negative sign ensures that higher values of $T_C$ correspond to smoother signals, in line with the sparsity measures introduced in Section~\ref{subsec:sparse}.
This formulation relies on two properties commonly satisfied by smoothing operators.

First, convergence to the mean:
\begin{align}
C_\alpha(I) \xrightarrow[\alpha \to \infty]{} \operatorname{mean}(I)
\end{align}
Second, monotonicity of the deviation with respect to the smoothing strength:
\begin{align}
|C_{\alpha_1}(I) - I| \ge |C_{\alpha_2}(I) - I| \quad \Leftrightarrow \quad \alpha_1 \ge \alpha_2
\end{align}
The exponential weighting prevents large deviations at high smoothing strengths from dominating the measure, while still accounting for the full range of smoothing scales.

In this work, we instantiate this framework using two smoothing operators:
\begin{itemize}
    \item Gaussian smoothing, where $\alpha = \sigma$ is the standard deviation of the Gaussian kernel, and
    \item Low-pass filtering, where $\alpha$ determines the cutoff frequency.
\end{itemize}
These choices allow us to capture smoothness both in the spatial domain and in the frequency domain.

\subsection{Taylor Approximation based Smoothness Measure}
We propose a smoothness measure based on first-order Taylor approximations.
This measure is motivated by the observation that smooth signals can be locally well approximated by their first-order Taylor expansion.
Let $I$ be a (possibly multi-dimensional) image function, let $a$ be a reference point, and let $\langle \cdot \, , \cdot \rangle$ be the inner product.
The first-order Taylor approximation $I^T_a$ of $I$ at $a$ is given by:
\begin{align}
    I^T_a(x) = I(a) + \langle \nabla I(a), (x-a) \rangle
\end{align}
For discrete images, numerical gradient approximations can be used to compute $\nabla I(a)$. To evaluate how well this local approximation represents the image, we consider a neighborhood $\mathcal{X}(x)$ for each pixel $x$, such as the 4-neighborhood or 8-neighborhood.
We then define an approximate image as:
\begin{align} \label{eq:ferror}
    I_\text{approx}(x) = \text{mean}_{a\in \mathcal{X}} I^T_a(x) 
\end{align}
Since both numerical gradient approximations and neighborhood averaging can be expressed as convolution operations, $I_{\mathrm{approx}}$ can be efficiently computed using a single convolution with a precomputed kernel.
If the first-order Taylor expansion provides a good local approximation, the difference $I_{\mathrm{approx}} - I$ should be close to zero.
We therefore define the Taylor-based smoothness measure with a distance measure $D$ as:
\begin{align}
    T_\nabla(I) = - D(I_\text{approx} - I)
\end{align}
Depending on the desired properties, any distance function $D$ can be applied here.
In this work, we choose the $\ell_2$ norm for $D$, but other norms or even the sparsity measures introduced in Section~\ref{subsec:sparse} are possible choices.
As before, the negative sign ensures that larger values correspond to smoother signals.

\section{Experimental Setup}\label{sec:experiments}
This section describes the experimental setup used in our study.
For reproducibility, all source code is available under \url{https://zivgitlab.uni-muenster.de/ag-pria/p-norms}.

\begin{figure}
    \begin{center}
    \begin{subfigure}{0.1\columnwidth}
        \scriptsize
        \begin{tabular}{l}
             1.00 \\ [31.5pt]
             1.01 \\ [31.5pt]
             1.10 \\ [31.5pt]
             1.20 \\ [31.5pt]
             1.40 \\ [31.5pt]
             1.60 \\ [31.5pt]
             1.80 \\ [31.5pt]
             2.00 \\ [15.725pt]
        \end{tabular}
        \caption*{$\ell_p$.}
    \end{subfigure}%
    \begin{subfigure}{0.18\columnwidth}
        \centering
        \includegraphics[width=.95\textwidth,trim={0 0 512 0},clip]{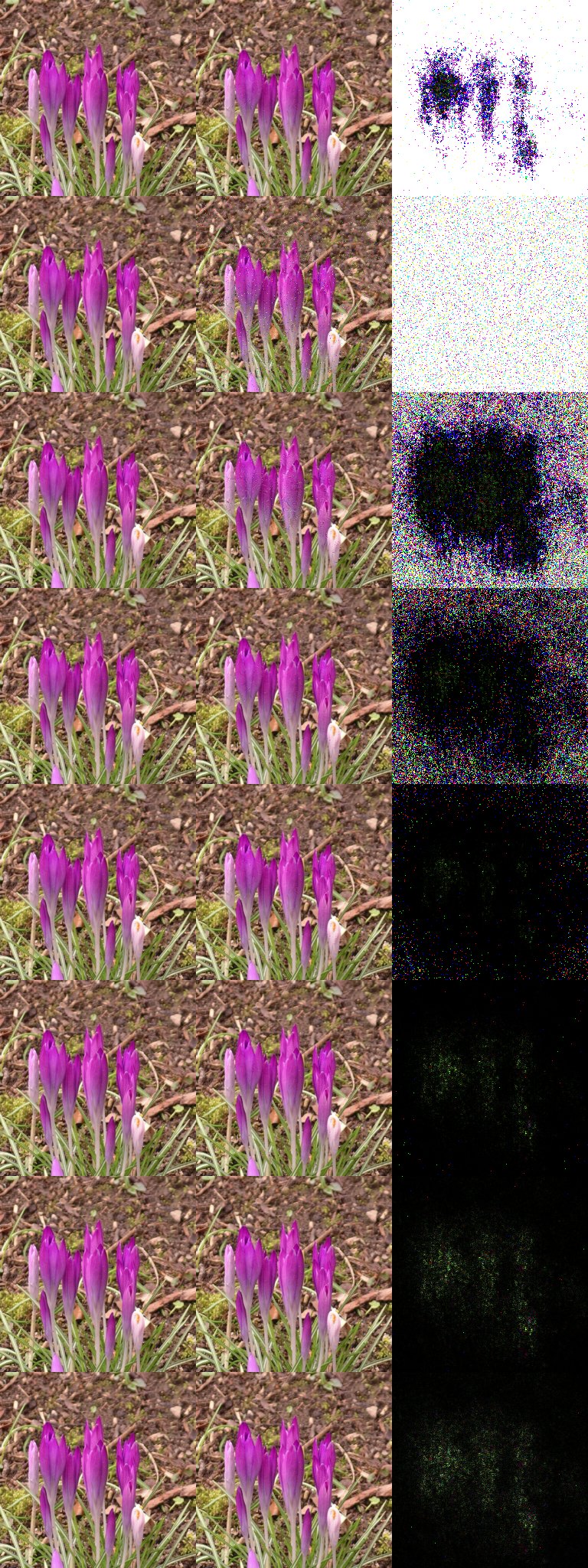}
        \caption*{Input.}
    \end{subfigure}%
    \begin{subfigure}{0.36\columnwidth}
        \centering
        \includegraphics[width=.95\textwidth,trim={256 0 0 0},clip]{imgs/90-resnet18-flowers102-46.png}
        \caption*{Normal training.}
    \end{subfigure}%
    \begin{subfigure}{0.36\columnwidth}
        \centering
        \includegraphics[width=.95\textwidth,trim={256 0 0 0},clip]{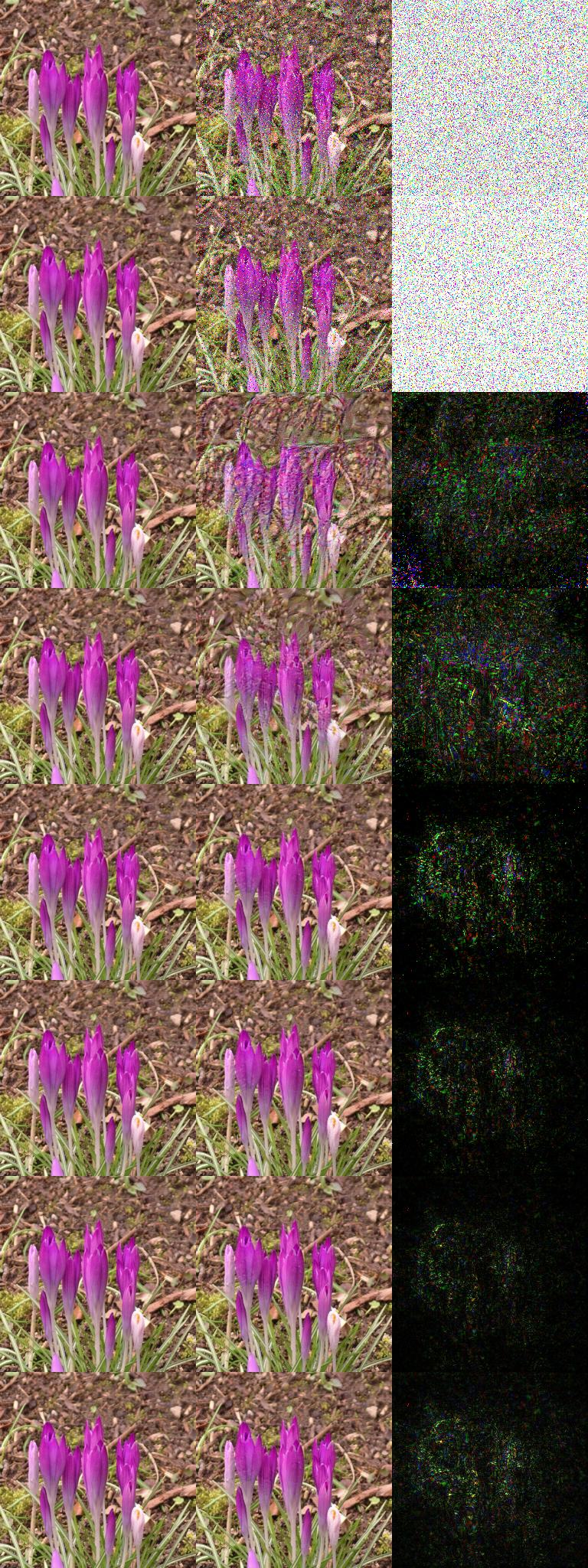}
        \caption*{Adversarial training.}
    \end{subfigure}
    \caption{Adversarially examples images for ResNet-18 and Flowers102. \textit{Columns (left to right):} $\ell_p$ norm constraint, original image, adversarial image, and perturbation $\delta$. \textit{Rows (top to bottom):} increasing values of $p \in \{1.00, 1.01, 1.10, 1.20, 1.40, 1.60, 1.80, 2.00\}$. White pixels in $\delta$ denotes zero-valued entries.}
    \label{fig:example-image}
    \end{center}
\end{figure}

\begin{figure*}
    \centering
    \includegraphics[width=0.7\textwidth]{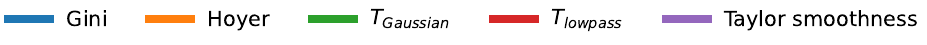}\\
    \begin{subfigure}{.47\textwidth}
        \centering
        \includegraphics[width=.95\columnwidth]{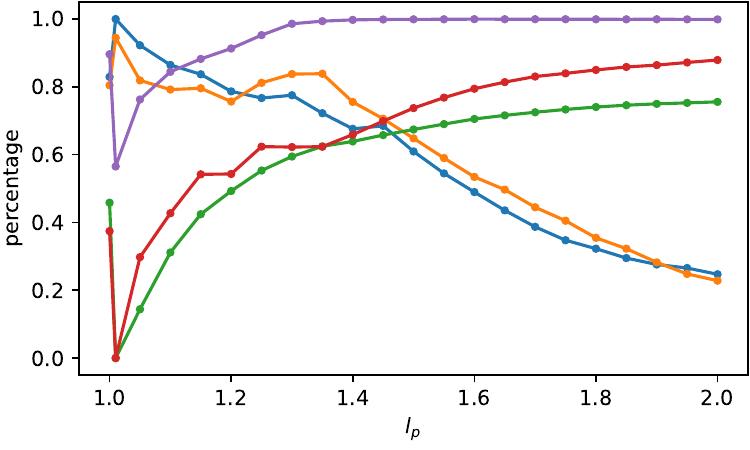}
        \caption{Attacks against normally trained models.}
        \label{fig:mean-measures-normal}
    \end{subfigure}%
    \begin{subfigure}{.47\textwidth}
        \centering
        \includegraphics[width=.95\columnwidth]{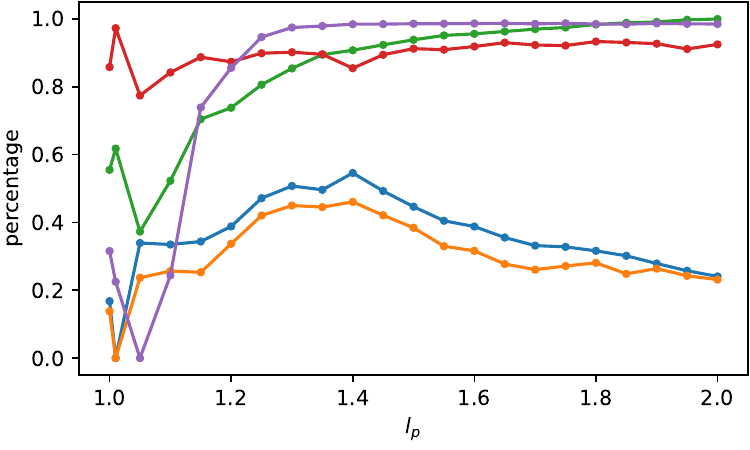}
        \caption{Attacks against adversarially trained models.}
        \label{fig:mean-measures-adv}
    \end{subfigure}
    \caption{Mean sparsity and smoothness as a function of $p$ averaged over all models and datasets. The curves are normalized independently per measure, model and dataset to the interval $[0,1]$ to enable comparison across measures. Sparsity measures are defined in Section~\ref{subsec:sparse} and smoothness measures in Section~\ref{sec:smooth}.}
    \label{fig:mean-measures}
\end{figure*}
\subsection{Models and Datasets}
We evaluate our approach on a diverse set of convolutional and transformer-based image classification models.
As convolutional architectures, we use ResNet \cite{he2016deep} variants ResNet-18, ResNet-50 and ResNet-101, VGG \cite{Simonyan2015very} variants VGG-16 and VGG-19. As transformer-based architectures, we use ViT-B/16 and ViT-B/32 from ViT \cite{Dosovitskiy2021an} and Swin-T v2 and Swin-S v2 from the Swin Transformer \cite{liu2021swin}.
The latter are comparable in computational complexity to ResNet-50 and ResNet-101, respectively, allowing for a meaningful comparison between convolutional and transformer-based models.

All models are evaluated on RGB image classification datasets with varying image resolutions and data distributions: CIFAR-10, CIFAR-100 \cite{krizhevsky2009learning}, Flowers102 \cite{nilsback2008automated} and \ac{gtsrb} \cite{houben2013detection}.

\subsection{Training Protocol}
All models are pretrained on ImageNet-1k \cite{deng2009imagenet} and subsequently fine-tuned on the target datasets.
Convolutional models are trained for 50 epochs, while transformer-based models are trained for 20 epochs.
We use the cross-entropy loss and the Adam optimizer, with learning rates of $10^{-3}$ for convolutional models and $5 \times 10^{-5}$ for transformer-based models.
For adversarial training, $75\%$ of each batch consists of adversarial examples. For each adversarial sample, the value of $p$ is sampled uniformly from the interval $p \in [1, 2]$.
Adversarial training on the pretrained models was performed for 20 epochs.
All experiments are performed on a server with AMD EPYC Milan 7543P CPU, 256 GB RAM and 2x NVIDIA A40 48 GB GPU. In total $\sim$2,000 GPU hours are used.

For the adversarial attacks, we use $p \in [1,2]$ with step size $0.05$.
As $\ell_1$-\ac{apgd} has a closed-form solution and \ac{afw} is an iterative algorithm, we also use $p=1.01$ to approximate the $\ell_1$ norm with \ac{afw}, since it is only defined for $p>1$ and even smaller $p$ result in numerical instability.

\subsection{Choice of Attack Magnitude \texorpdfstring{$\varepsilon$}{eps}}
The choice of the attack magnitude $\varepsilon$ is often made arbitrarily in the literature and typically justified through qualitative inspection of adversarial examples.
This makes quantitative comparisons across models, datasets, and norm choices difficult.
We value a consistent choice of $\varepsilon$ over visually good looking adversarial attacks to decrease variability and increase comparability between $\ell_p$ norms, models and datasets.

To ensure comparability, we adopt a performance-based strategy to determine $\varepsilon$ for each model, dataset, and norm and denote it by $\varepsilon_p$.
We do this as follows: 
\begin{enumerate}
    \item Test the performance of each model on each dataset.
    \item For every combination of model \& dataset, set the target accuracy as $\frac{1}{3}$ of the test accuracy.
    \item For each model \& dataset choose for every $\ell_p$ norm the $\varepsilon_p$, so that the test accuracy degrades to that target accuracy (but not further).
\end{enumerate}
This procedure ensures that all attacks achieve comparable effectiveness, allowing us to isolate the influence of the norm choice on sparsity and smoothness.

\begin{figure*}
    \begin{subfigure}{.47\textwidth}
        \centering
        \includegraphics[width=.95\columnwidth]{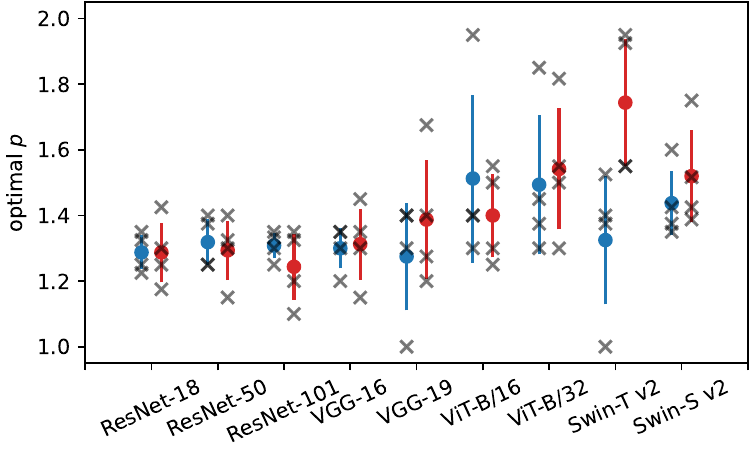}
        \caption{Optimal $p$ per model aggregated over datasets.}
        \label{fig:best-p-models}
    \end{subfigure}%
    \begin{subfigure}{.47\textwidth}
        \centering
        \includegraphics[width=.95\columnwidth]{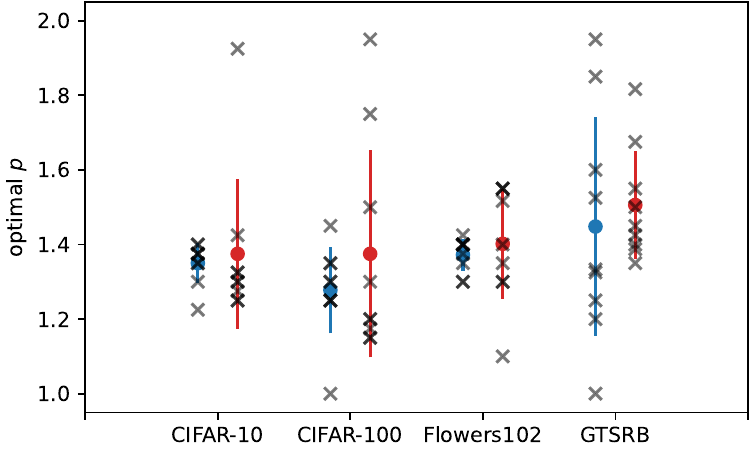}
        \caption{Optimal $p$ per dataset aggregated over models.}
        \label{fig:best-p-datasets}
    \end{subfigure}
    \caption{Optimal values of $p$ maximizing sparsity and smoothness. Dots indicate the mean optimal value and error bars the standard deviation; individual runs are shown as crosses. Blue denotes attacks against normally trained models and red denotes attacks against adversarially trained models.}
    \label{fig:best-p}
\end{figure*}

\subsection{Calculating Optimal \texorpdfstring{$p$}{p}}
Our goal is to identify values of $p$ for which adversarial attacks simultaneously maximize sparsity and smoothness.
Let $\mathcal{M}_S$ denote the set of all sparsity measures, $\mathcal{M}_T$ the set of all smoothness measures, and $\mathcal{M} = \mathcal{M}_S \cup \mathcal{M}_T$ their union and thus the set of all measures we are interested in.
Each measure $m \in \mathcal{M}$ is normalized to the interval $[0, 1]$.
We first compute the largest value $\beta_{\text{opt}} \ge 0$ such that there exists at least one value of $p$ for which all measures exceed $\beta_{\text{opt}}$:
\begin{align}\label{eq:beta_opt}
    \beta_\text{opt} = \max \{\beta \geq 0 \ | \underset{m \in \mathcal{M}}{\cap} \{ p: m(p) \geq \beta \} \neq \emptyset \}
\end{align}
Using $\beta_{\text{opt}}$, we define the set of optimal norm parameters as:
\begin{align}
    \mathcal{M}_{p_\text{opt}} = \underset{m \in \mathcal{M}}{\cap} \{ p: m(p) \geq \beta_\text{opt} \}
\end{align}
This construction guarantees that $\mathcal{M}_{\text{opt}}$ is non-empty and yields either a single optimal value or a set of equally optimal values of $p$.
At the same time, $\beta_{\text{opt}}$ provides a quantitative measure of how sparse and smooth the attacks are for the given model and dataset combination.

\begin{figure*}
    \begin{subfigure}{.47\textwidth}
        \centering
        \includegraphics[width=.95\columnwidth]{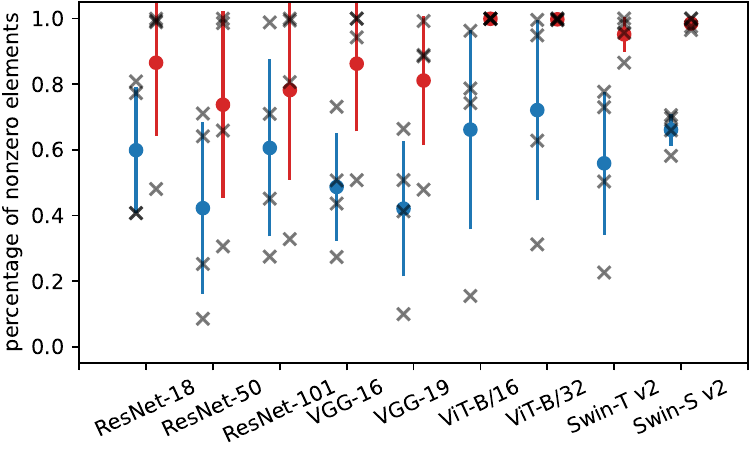}
        \caption{Attacks generated using $\ell_1$-\ac{apgd}.}
    \end{subfigure}%
    \begin{subfigure}{.47\textwidth}
        \centering
        \includegraphics[width=.95\columnwidth]{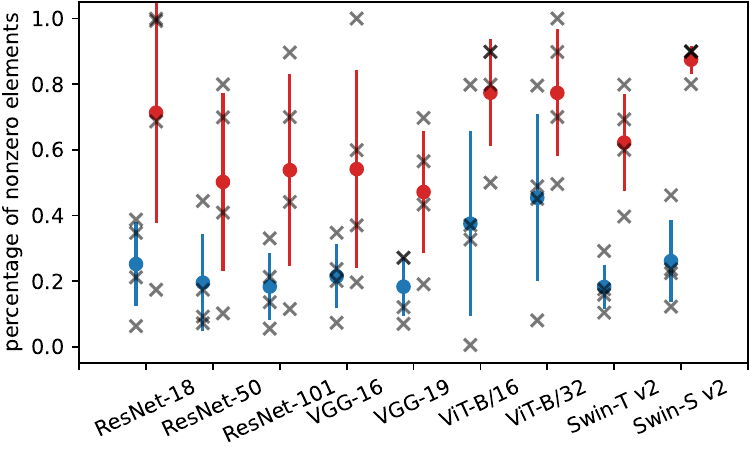}
        \caption{Attacks generated using $\ell_{1.01}$-\ac{afw}.}
    \end{subfigure}
    \caption{$\ell_0$ sparsity of adversarial perturbations $\delta$ normalized by image size, shown as percentage of non-zero pixels. Mean and std given by dot and error bar, individual data points are represented as crosses. Blue denotes attacks against normally trained models and red denotes attacks against adversarially trained models.}
    \label{fig:l0-apgd-afw}
\end{figure*}

\subsection{\texorpdfstring{$\alpha$}{alpha} Schedules for Smoothing Operation based Smoothness Measures}
For the smoothness operation based smoothness measures $T_{C_\alpha}$ we use the following $\alpha$ schedules.
For Gaussian smoothing we set $\sigma_i^2 = \alpha_i^2$ and linearly sample $\alpha_i \in [1, 10]$ with step size $1$.
For the low pass filter, we use the same $\alpha_i \in [1, 10]$ and set the cutoff frequencies $c_i$ as $c_i = (10 - \alpha_i + 1)^2$ linearly scaled to half the image size.
This ensures big cutoff steps for small $\alpha_i$ meaning high frequencies with lower impact on smoothing get discarded quickly.
Also, the smaller steps for higher $\alpha_i$ ensure a good sampling of the stronger smoothed images with high impact $\alpha_i$.

\section{Results}\label{sec:results}
In this section, we present an extensive empirical evaluation of the influence of the choice of the $\ell_p$ norm on the sparsity and smoothness of adversarial attacks.
Unless stated otherwise, all reported results are averaged over multiple models and datasets.

\subsection{Sparsity and Smoothness as a Function of \texorpdfstring{$p$}{p}}
We now analyze how sparsity and smoothness evolve as a function of the norm parameter $p \in [1, 2]$.
Across all models and datasets, we observe that smoothness increases with increasing $p$, while sparsity generally decreases as shown in Fig.~\ref{fig:mean-measures}.
However, neither trend is linear, and sparsity in particular exhibits non-monotonic behavior.
Most notably, both sparsity measures have a peak around $p=1.3$ and for adversarial training even attain their maximum.
This indicates that sparsity is not maximized at $p = 1$, despite the common association between the $\ell_1$ norm and sparse solutions.
Especially for low $p$ and after adversarial training, all measures do not increase/decrease monotonously.
It stands out that between $p=1$ and $p=1.01$ the behavior of both sparsity and smoothness measures is strongly influenced by the attack algorithm changing from $\ell_1$-\ac{apgd} to \ac{afw} as the attack algorithm.
Although both are based on \ac{apgd}, the closed-form solution of $\ell_1$-\ac{apgd} leads to noticeably less sparse and smoother perturbations than the \ac{afw}-generated approximations.
To further analyze this phenomenon, we conducted a separate study on these two versions (Section~\ref{subsec:l1-apgd-afw}).
The measures on the \ac{afw}-generated adversarial attacks converge more evenly for $p \rightarrow 1$ after the normal training than after adversarial training.

\subsection{Optimal \texorpdfstring{$p$}{p} across Models and Datasets}
When aggregating optimal values of $p$ across datasets for each model (see Fig.~\ref{fig:best-p-models}), convolutional architectures exhibit a tight clustering of optimal values around $p=1.3$.
This behavior is consistent across both normally trained and adversarially trained models.
In contrast, transformer-based architectures show greater variability.
Their optimal values of $p$ are typically higher, often in the range $p \in [1.4, 1.5]$ with larger standard deviations.
An exception is the Swin-T v2 model which has an average optimal $p$ of $1.3$ for normal training and $1.7$ for adversarial training.

When aggregating across models for each dataset in Fig.~\ref{fig:best-p-datasets}, we observe lower optimal $p$ and standard deviation for the normal training than after adversarial training.
Only the \ac{gtsrb} has a higher standard deviation for normal training.
The outliers with high $p$ are from ViT-B/16 and ViT-B/32.

The transformer-based models, especially the ViT, are inherently more robust against adversarial attacks with small $p$.
Aside from the higher optimal $p$, Fig.~\ref{fig:l0-apgd-afw} shows that the adversarial attacks for $p$ near $1$ do use a high percentage of all available pixels.
For the adversarial trained models, almost all of the image has to be attacked.

\subsection{Comparison of \texorpdfstring{$\ell_1$}{l1}-APGD and \texorpdfstring{$\ell_{1.01}$}{l1.01}-AFW} \label{subsec:l1-apgd-afw}
Since Fig.~\ref{fig:mean-measures} shows that adversarial attacks generated with $\ell_1$-\ac{apgd} are more sparse than those generated with \ac{afw} using $p=1.01$, we further analyze the sparsity behavior of these two attack methods using the stricter $\ell_0$ measure across all evaluated models.
As shown in Fig.~\ref{fig:l0-apgd-afw}, $\ell_1$-\ac{apgd} produces less sparse adversarial attacks than \ac{afw} for most models, regardless of whether the models are normally trained or adversarially trained. 
Note that for $\ell_0$ low scores mean a more sparse result.
This effect is consistent across datasets and architectures.
In addition, Fig.~\ref{fig:l0-apgd-afw} shows that adversarial training generally leads to less sparse attacks for both methods, indicating that adversarial training increases robustness particularly against sparse perturbations.

Qualitative inspection of the adversarial examples in Fig.~\ref{fig:example-image} further supports these findings.
Before adversarial training, $\ell_1$-\ac{apgd} generates perturbations that are more concentrated on salient object regions than those generated by \ac{afw}, even for values of $p$ close to $1$.
However, this distinction largely vanishes after adversarial training.

Overall, these results indicate that the observed differences in sparsity between $\ell_1$-\ac{apgd} and \ac{afw} are not solely due to the choice of the norm parameter $p$, but are also influenced by the specific optimization algorithm used to generate the adversarial attacks.

\subsection{Discussion}
As expected, sparsity increases and smoothness decreases for $p\to 1$.
Notably however, this behavior is neither linear nor monotonic over the interval $p \in [1,2]$.
In particular, sparsity does not continuously increase toward $p = 1$, but instead reaches a maximum around $p=1.3$.
The main reason for this is the different behaviors of $\ell_1$-\ac{apgd} compared to the approximative $\ell_1.01$-\ac{afw}, as further analyzed in Section~\ref{subsec:l1-apgd-afw}.
As for smoothness measures, all evaluated smoothness measures indicate that smoothness plateaus at around $p \approx 1.4$.
Increasing $p$ further does not lead to smoother attacks, but does lead to less sparse attacks.
In particular, the interval $p \in [1.4, 1.8]$ shows the steepest decrease in sparsity.
This behavior indicates the existence of a 'sweet spot' for the choice of $p$ within the interval $[1.3, 1.5]$, which is consistent with the optimal values observed in Fig.~\ref{fig:best-p}.

Fig.~\ref{fig:best-p} further shows that adversarial training leads to higher variance and generally larger optimal values of $p$, particularly for transformer-based models.
In addition, Fig.~\ref{fig:l0-apgd-afw} indicates that adversarial attacks with values of $p$ close to $1$ require modifying a large fraction of the available pixels after adversarial training.
These observations suggest that adversarial training is particularly effective at improving robustness against sparse adversarial attacks.

\section{Conclusion}\label{sec:conclusion}
In this work, we investigated the effect of the choice of the $\ell_p$ norm on the properties of adversarial attacks.
In particular, we analyzed how sparsity and smoothness behave when adversarial attacks are constructed under $\ell_p$ norm restrictions for $p \in [1, 2]$.
To quantify sparsity, we employed two established sparsity measures that are well suited for fixed-size inputs such as images.
To quantify smoothness, we introduced a general framework for deriving smoothness measures based on smoothing operations and proposed two instantiations of this framework.
In addition, we introduced a smoothness measure based on first-order Taylor approximations.

Using these measures, we conducted a comprehensive empirical evaluation across multiple datasets and a diverse set of model architectures, including both convolutional and transformer-based networks.
Our results show that both sparsity and smoothness are strongly influenced by the choice of the norm parameter $p$.
In particular, we find that the commonly used choices $p=1$ and $p=2$ are generally suboptimal when sparsity and smoothness are considered jointly.
Instead, intermediate values of $p$ consistently yield adversarial perturbations with more favorable trade-offs between these two properties.

Moreover, we show that the optimal choice of $p$ depends on both the model architecture and the data distribution.
This dependency is especially pronounced for transformer-based models, which exhibit higher optimal values of $p$ and greater variability compared to convolutional architectures which tend to agree on an optimal value of $p \approx 1.3$.

Finally, adversarial training substantially reduces the effectiveness of sparse attacks, forcing successful perturbations to modify a larger fraction of the input.

Overall, our findings show that the commonly used practice of selecting the $\ell_p$ norm without further justification is suboptimal.
Careful selection of the norm parameter $p$ enables better control over sparsity and smoothness and leads to more informative and comparable adversarial evaluations.
We believe that these insights provide a useful foundation for future research on perceptually grounded adversarial attacks.

\section*{Acknowledgment}
ChatGPT 5.2 was used exclusively for sentence-level reformulation, but not for text generation.

\bibliographystyle{IEEEtran}
\bibliography{bibliography}
\end{document}